\documentclass[10pt,twocolumn,letterpaper]{article}

\usepackage{cvpr}              

\usepackage[dvipsnames]{xcolor}

\definecolor{darkred}{rgb}{0.7, 0.0, 0.0}
\definecolor{darkgreen}{rgb}{0.0, 0.37, 0.14}
\definecolor{darkblue}{rgb}{0.10, 0.17, 0.8}

\newcommand{\model}{Obj-NeRF\xspace}

\definecolor{cvprblue}{rgb}{0.21,0.49,0.74}
\usepackage[pagebackref,breaklinks,colorlinks,citecolor=cvprblue]{hyperref}
\usepackage{bm}
\usepackage{bbm}
\usepackage{algorithm} 
\usepackage{algorithmic} 
\usepackage{setspace}
\usepackage{float}

\newcommand{\RNum}[1]{\uppercase\expandafter{\romannumeral #1\relax}}

\def\paperID{11400} 
\def\confName{CVPR}
\def\confYear{2024}

\title{Obj-NeRF: Extract Object NeRFs from Multi-view Images}

\author{
    Zhiyi Li$^1$, Lihe Ding$^2$, and Tianfan Xue$^2$ \\
    $^1$Tsinghua University, $^2$The Chinese University of Hong Kong \\
    {\tt\small lizhiyi20@mails.tsinghua.edu.cn, \{dl023,tfxue\}@ie.cuhk.edu.hk}
}

\begin{document}

\twocolumn[{
	\renewcommand\twocolumn[1][]{#1}
	\maketitle
	\thispagestyle{empty}
	\vspace{-30pt}
	\begin{center}
            \includegraphics[width=\linewidth]{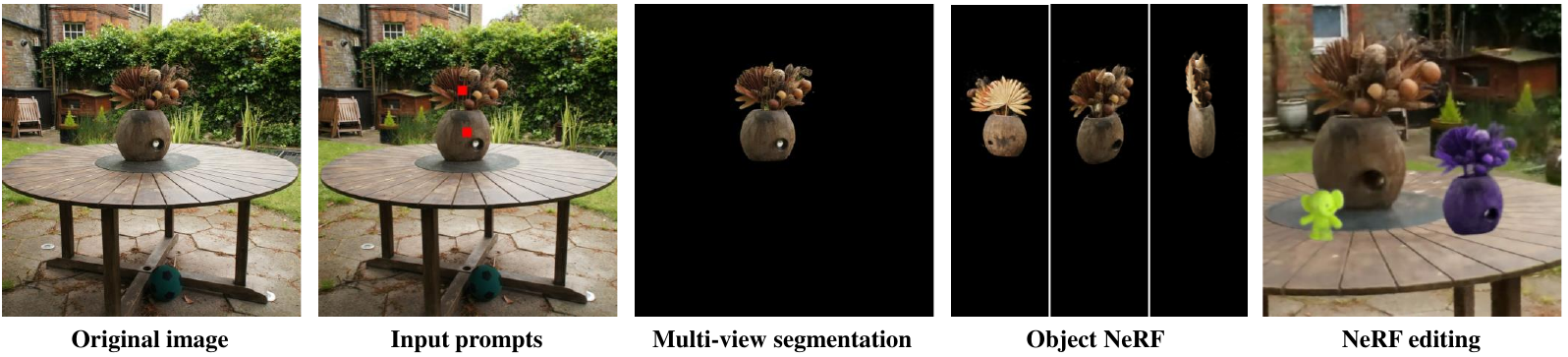}
            \vspace{-20pt}
            \captionof{figure}{
            Proposed Obj-NeRF: indicate prompts on an image, then Obj-NeRF will output the segmented NeRF for the target object. With the segmented object NeRF, some applications including NeRF editing can be realized.
            }
		\label{fig:intro}
		\vspace{2pt}
	\end{center}
}]

\begin{abstract}
\vspace{-5pt}
Neural Radiance Fields (NeRFs) have demonstrated remarkable effectiveness in novel view synthesis within 3D environments. However, extracting a radiance field of one specific object from multi-view images encounters substantial challenges due to occlusion and background complexity, thereby presenting difficulties in downstream applications such as NeRF editing and 3D mesh extraction. To solve this problem, in this paper, we propose Obj-NeRF, a comprehensive pipeline that recovers the 3D geometry of a specific object from multi-view images using a single prompt. This method combines the 2D segmentation capabilities of the Segment Anything Model (SAM) in conjunction with the 3D reconstruction ability of NeRF. Specifically, we first obtain multi-view segmentation for the indicated object using SAM with a single prompt. Then, we use the segmentation images to supervise NeRF construction, integrating several effective techniques. Additionally, we construct a large object-level NeRF dataset containing diverse objects, which can be useful in various downstream tasks. To demonstrate the practicality of our method, we also apply Obj-NeRF to various applications, including object removal, rotation, replacement, and recoloring. The project page is at \href{https://objnerf.github.io/}{https://objnerf.github.io/}. 
\end{abstract}    
\section{Introduction}
\label{Introduction}

Neural Radiance Fields (NeRFs) have attracted enormous academic interest in 3D scene representation, due to their remarkable ability to produce high-quality synthesized views in diverse 3D environments~\cite{mildenhall2021nerf}. Recent works have concentrated on enhancing the performance and practicality of NeRF, thereby broadening its applicability with higher reconstruction quality and faster training speed~\cite{sun2022direct, deng2022depth, muller2022instant}. 

Moreover, NeRF is also widely used in many downstream applications, including 3D editing and novel view synthesis~\cite{liu2023zero}. With that, the demand for object-specific NeRF datasets has risen. Nevertheless, the inherent limitation of NeRF, which provides only color and density information, presents a challenge for extracting specific objects from multi-view images. 

\begin{figure*}[ht]
    \centering
    \includegraphics[width=\linewidth]{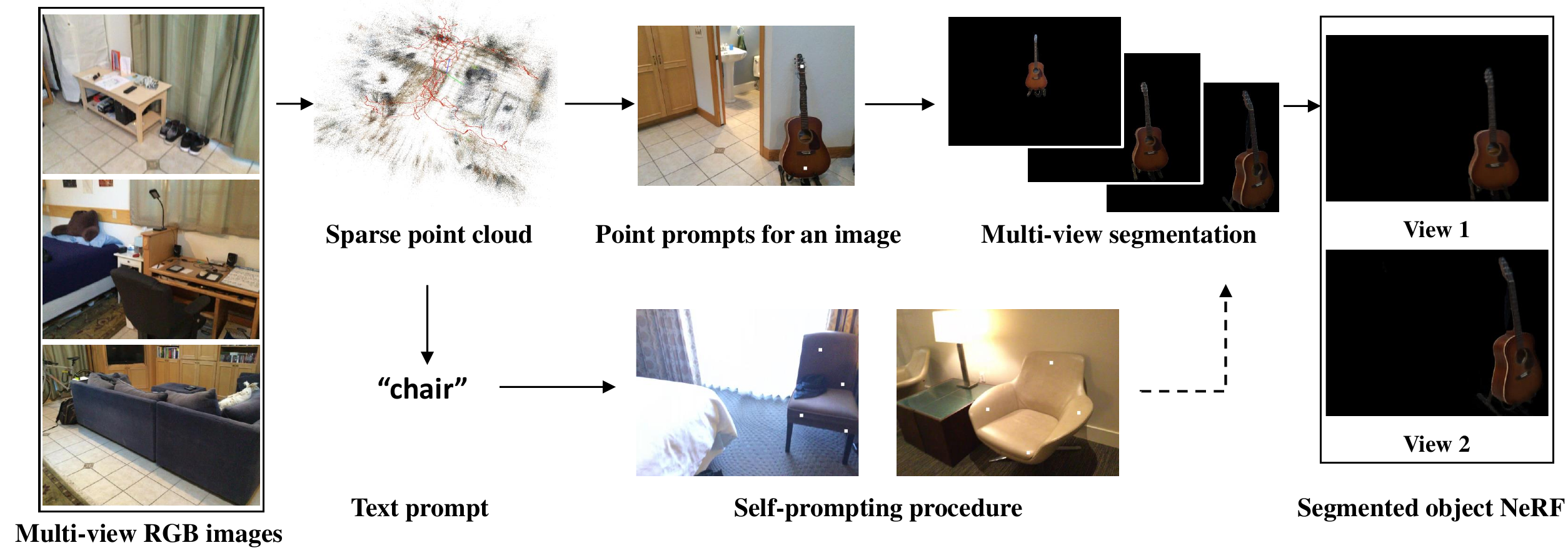}
    \caption{The overall pipeline for Obj-NeRF. Starting with multi-view RGB images, a COLMAP sparse point cloud can be constructed, which provides multi-view consistency for segmentation. After initializing several prompts for the first image, we can automatically obtain multi-view segmented images quickly, which are used to construct the required segmented NeRF. For large datasets, the indicated objects will be prompted for each scene.}
    \label{fig:overall pipeline}
\end{figure*}

In order to extract object-level NeRF from multi-view images, recent works have primarily explored the utilization of 2D visual models like CLIP~\cite{radford2021learning} or DINO~\cite{caron2021emerging} along with additional feature images provided by modified NeRFs, such as LERF~\cite{kerr2023lerf} and Interactive Segment Anything NeRF~\cite{chen2023interactive}. However, this approach has limitations, including the absence of required 3D object meshes, excessive additional training costs for the complete scene NeRF, and suboptimal reconstruction quality~\cite{chen2023interactive}. 

To address these limitations, our objective is to extract specific object NeRFs from multi-view images representing a 3D scenario. Although there have been continuous advancements in 2D image segmentation, such as the recently proposed Segment Anything Model (SAM)~\cite{kirillov2023segment}, segmenting the required 3D object NeRF from an original scenario encounters numerous challenges. Notably, training a 3D segmentation model similar to SAM for zero-shot segmentation tasks remains a formidable undertaking~\cite{cen2023segment}. While extending segmentation capabilities directly into 3D scenarios presents formidable challenges, combining the 2D segmentation proficiency of SAM with the 3D representation capabilities of NeRF is a feasible and promising approach. 

Based on this idea, in this work, we propose Segment Object NeRF (Obj-NeRF) to extract a certain object and reconstruct its geometry, from a few user prompts. The effectiveness of Obj-NeRF is depicted in Figure~\ref{fig:intro}. Obj-NeRF initially receives prompts indicating a specific object within a single image, selected from a set of multi-view images representing a 3D scenario. Subsequently, Obj-NeRF generates multi-view segmented images through the utilization of the SAM, thereby supervising the construction of the segmented target object NeRF. After acquiring the object NeRF, we further evaluate it on various 3D editing applications. The main contributions of this paper can be summarized as follows: 
\begin{itemize}
    \item Firstly, we present a comprehensive pipeline for constructing a segmented NeRF targeting a specific object, with the input consisting of initial prompts from a single image. Our pipeline eliminates the need for pre-trained full-scene NeRFs, thereby avoiding unnecessary training expenses and enhancing reconstruction quality. 
    \item Next, to obtain multi-view segmented images of the target object, we introduce a multi-view segmentation algorithm with high quality and efficiency. Leveraging a 3D sparse point cloud, we rapidly disseminate the initial prompts to all images and extract segmented masks from SAM. 
    \item Additionally, in order to create massive object NeRFs from large multiview datasets, we propose an automatic self-prompting mechanism with only simple textual input. It will enable the identification of the desired object for each scene, thereby constructing a dataset of multi-view segmented objects. 
    \item Finally, to enhance the quality of novel view synthesis using NeRFs, we propose several methods, including supervision with sparse and dense depth priors, bounding box calculation, and ray pruning for improving performance. Additionally, we validate the effectiveness of segmented object NeRFs by modifying existing NeRFs. 
\end{itemize}

\section{Related Works}
\label{Related Works}

\textbf{NeRF for Novel View Synthesis.}
Neural Radiance Fields (NeRFs) have gained numerous research interests in novel view reconstruction~\cite{mildenhall2021nerf}. Recently, researchers have been working on improving the effectiveness of NeRFs, including enhancing reconstruction quality via various methods~\cite{deng2022depth, roessle2022dense, wang2022nerf, barron2021mip}, increasing training speed with high synthesis effect~\cite{sun2022direct, tancik2023nerfstudio, muller2022instant}, and expanding their application scenarios~\cite{pumarola2021d, tang2022compressible}. There are also many downstream works on NeRFs, such as NeRF editing~\cite{yuan2022nerf, yin2023or, mirzaei2023spin}, 3D mesh extraction~\cite{munkberg2022extracting}, and 3D generation tasks~\cite{lin2023magic3d, liu2023zero, poole2022dreamfusion}.  

\textbf{Segmentation on NeRFs.} 
Significant progress has been made in 2D semantic fields, including DETR~\cite{carion2020end} and CLIP~\cite{jia2021scaling}. Recently, models trained on extremely large-scale datasets, such as the Segment Anything Model (SAM)~\cite{kirillov2023segment} and SEEM~\cite{zou2023segment}, have shown strong ability on zero-shot image segmentation. Based on these, many researchers have made some progress to expand on 3D segmentation fields, by training an extra semantic feature on modified NeRF~\cite{kerr2023lerf} and distilling segmentation backbone with NeRF~\cite{chen2023interactive}. However, these works cannot provide a segmented object NeRF, which is essential in many downstream applications like 3D scenario editing. SA3D~\cite{cen2023segment} has proposed a method to construct a segmented object NeRF from multi-view images with SAM. Nonetheless, SA3D requires a pre-trained original full-scene NeRF, which is impractical and brings extra training costs and low reconstruction quality, especially for large-scale scenes.

\section{Preliminaries} 
\label{Preliminaries} 

In this section, some preliminaries for Obj-NeRF will be introduced briefly, including the Neural Radiance Fields (NeRFs)~\cite{mildenhall2021nerf} and the Segment Anything Model (SAM)~\cite{kirillov2023segment}. 

\paragraph{Neural Radiance Field.} NeRF presents an effective way to synthesize novel views in 3D scenarios. 
Specifically, NeRF defines an underlying continuous volumetric scene function $\mathcal{F}_{\bm{\Theta}}: (\bm{x}, \bm{d}) \Longrightarrow (\bm{c}, \sigma)$, which outputs the color $\bm{c} \in \mathbb{R}^3$ and the volume density $\sigma \in \mathbb{R}^+$ with a given spacial location $\bm{x} \in \mathbb{R}^3$ and viewing direction $\bm{\theta} \in \mathbb{S}^2$. 
In this way, the rendering color $C(\bm{r})$ for a specific camera ray $\bm{r}(t) = \bm{o} + t \bm{d}$ can be expressed by a volume rendering algorithm as follows: 
\begin{equation}
    C(\bm{r}) = \int_{t_n}^{t_f} T(t) \sigma(\bm{r}(t)) \bm{c}(\bm{r} (t), \bm{d}) \,dt, 
\end{equation}
where $t_n$ and $t_f$ are the near and far bounds, and the accumulated transmittance $T(t)$ can be calculated as: 
\begin{equation}
    T(t) = \exp{-\int_{t_n}^t \sigma(\bm{r} (s)) \,ds}. 
\end{equation}
With these definitions, NeRFs can be optimized using the loss between the ground-truth color $C(\bm{r})$ and the calculated color $\hat{C} (\bm{r})$ for any image $I$: 
\begin{equation}
    \mathcal{L}_I = \sum_{\bm{r} \in I} \left\Vert C(\bm{r}) - \hat{C}(\bm{r}) \right\Vert^2. 
\end{equation}


\paragraph{Segment Anything Model (SAM)} SAM, training by numerous 2D images, has been proved to achieve a state-of-art efficiency in zero-shot segmentation tasks~\cite{kirillov2023segment}. With an image $\bm{I}$ and some prompts $\mathcal{P}$, including points (positive or negative) and boxes, SAM can provide a mask for the indicated object $\text{mask} = \mathcal{S}(\bm{I}, \mathcal{P})$.

\section{Methods}
\label{Methods}

In this section, we will introduce details of \model. First, the overall pipeline will be demonstrated in \cref{Overall Pipeline}. Then, a one-shot multi-view segmentation method will be presented in \cref{Multi-view Segmentation}. After that, we will introduce a self-prompting method to construct an object NeRF dataset including massive objects using the segmentation method above in \cref{Large Dataset Self-prompting}. In the end, some strategies for novel views synthesizing by NeRFs will be provided in \cref{Novel View Synthesizing}. 

\subsection{Overall Pipeline}
\label{Overall Pipeline}

We consider a set of multi-view images $\bm{I} = {I_1, ..., I_n}$ for one specific scenario with known camera poses. If not, structure-from-motion methods like COLMAP~\cite{schonberger2016structure} can be utilized to estimate them. The objective is to acquire the 3D representation for any object segmented from this scenario with few prompts. To achieve this, a pre-trained full-scene NeRF for the scenario is not required due to the unnecessary training cost and relatively poor quality. Thus, we propose a method to acquire the segmented NeRF for the object from multi-view images directly. 

The overall pipeline is shown in Fig.~\ref{fig:overall pipeline}.  To start with, users will first provide a few prompts for one image on the object that is expected to be segmented. Based on this, a COLMAP sparse point cloud~\cite{schonberger2016structure} can be constructed, which provides the correspondence between 2D images and point clouds used in the next step. Then, the multi-view segmentation procedure with SAM will provide multi-view masks for the object. For large datasets, such as ScanNet~\cite{dai2017scannet}, ScanNet++~\cite{yeshwanth2023scannet++}, and 3RScan~\cite{wald2019rio}, the self-prompting procedure will quickly generate a series of prompts of a kind of objects, which can be used to acquire multi-view segmented for each scene with the method above. In the end, the segmented NeRF of the indicated object will be trained with these multi-view segmented images, which provides novel view synthesizing abilities. 

\subsection{Multi-view Segmentation}
\label{Multi-view Segmentation} 

\subsubsection{Multi-view Segmentation Algorithm}

The first step is quickly obtaining multi-view segmented images from initial prompts on one specific image. It is easy to get the mask for the initial image $M_0$ with SAM. However, multi-view consistency should be utilized here in order to segment the indicated object on each image. A similar approach is also used by Yin et al.~\cite{yin2023or} to find a 2D-3D geometry match relationship with prompts spreading, but here we use it for a different task with some effective methods mentioned as follows. 

More specifically, a sparse point cloud can be easily constructed from input images using COLMAP, a 3D reconstruction toolbox~\cite{schonberger2016structure}. These sparse point clouds provide the correspondence between feature points on each image and the 3D points in the point cloud. In this way, we can construct a 3D point list $\bm{D}$, which contains 3D points belonging to the indicated object. After initializing the list with 3D points that correspond to the feature points on $I_0[M_0]$, the 3D point list and the masks of remnant images can be updated iteratively. Specifically, for a new image $I_i$, the point prompts $\bm{p}_i$ can be selected from the feature points that correspond to 3D points in the list. Then, the mask $M_i$ can be obtained using SAM segmentation model $\mathcal{S}(I_i, \bm{p}_i)$. 
After that, all feature points on $I_i[M_i]$ can be added to the list $\bm{D}$, which finishes an iterating step. The multi-view segmentation procedure can be summarized in \cref{alg:Multi-view Segmentation}.

\begin{algorithm}[t]
    \caption{Multi-view Segmentation} 
    \label{alg:Multi-view Segmentation} 
    \setstretch{1.35}
    \begin{algorithmic}[1]
        \REQUIRE A set of images $I_0, I_1, I_2, ..., I_n$; Initial point prompts $\bm{p}_0$; SAM model $\mathcal{S}$; COLMAP sparse point cloud $\mathcal{C}$. 
        \ENSURE Multi-view segmented masks $M_1, M_2, ..., M_n$. 
        \STATE Get the mask for the initial image $M_0 = \mathcal{S}(I_0, \bm{p}_0)$ 
        \STATE Find all feature points $\bm{X} =  \mathcal{C}[I_0] \cup I_0[M_0]$ 
        \STATE Init 3D point list $\bm{D} = \mathcal{C}(\bm{X}) $. 
        \FOR {$i = 1, 2, ..., n$} 
            \STATE Find point prompts $\bm{p}_i$ for $I_i$ from $\bm{D}$ and $\mathcal{C}$ 
            \STATE $M_i \leftarrow \mathcal{S}(I_i, \bm{p}_i)$ 
            \STATE $\bm{X}_i \leftarrow \mathcal{C}[I_i] \cup I_i[M_i]$ 
            \STATE $\bm{D} \leftarrow \bm{D} \cup \mathcal{C}(\bm{X}_i)$
        \ENDFOR
        \RETURN $M_1, M_2, ..., M_n$
    \end{algorithmic}
\end{algorithm}

\begin{figure}[t]
    \centering
    \includegraphics[width=\linewidth]{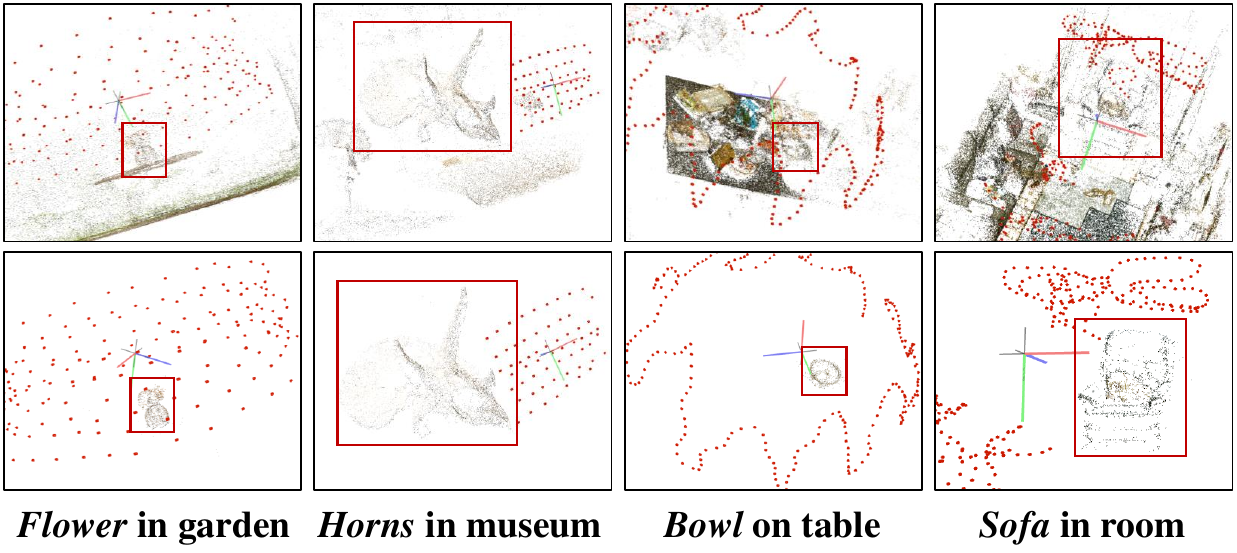}
    \caption{Segmented point cloud of target objects; Images in the first row are original sparse point clouds; Images in the second row are segmented point clouds; Red points on these images indicate the position of cameras.}
    \label{fig:segmented point cloud}
\end{figure}

After executing the algorithm above, an interesting byproduct will be obtained from the list $\bm{D}$. As the definition of $\bm{D}$, it is consisted of those 3D points belonging to the indicated object, which provides its sparse point cloud. Fig.~\ref{fig:segmented point cloud} shows the segmented sparse point cloud of these indicated objects. This can be used in the next step and will offer some priors to the novel views synthesizing procedure in Subsection \ref{Novel View Synthesizing}. 

\begin{figure}[t]
    \centering
    \includegraphics[width=\linewidth]{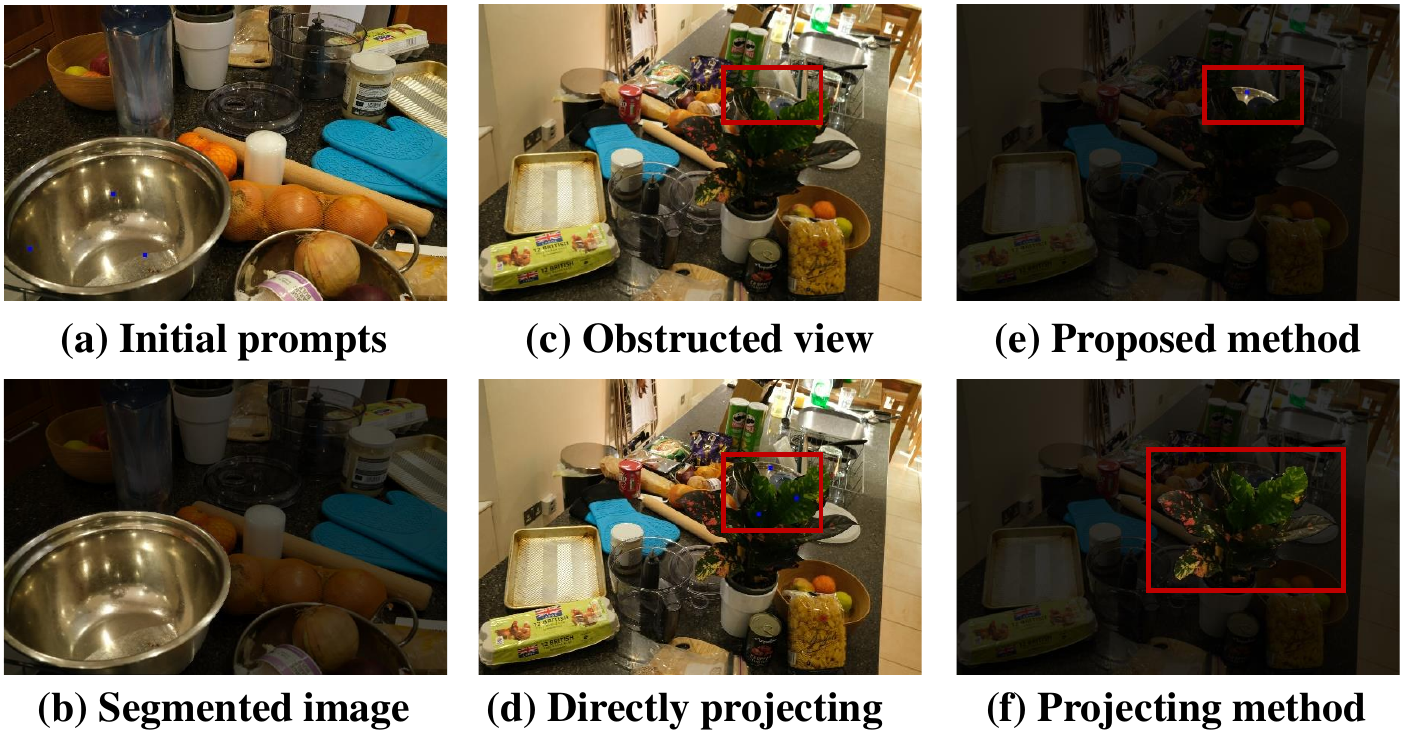}
    \caption{Showing the obstruction effect;
    (a) Initial point prompts for the target bowl; (b) Segmented image for the first image; (c) A view where the target bowl is obstructed by the plant; (d) and (f) Directly projecting method leading to a wrongly segmented object; (e) Proposed method to get the correct segmentation.
    }
    \label{fig:obstruction}
\end{figure}

\subsubsection{Obstruction Handling}

There is a thorny problem that when the target object is obstructed by other things, it will be easy for the procedure leading to a wrong result. In \cite{yin2023or}, the method based on projecting 3D points directly as the prompts will severely suffer from it. Here, our proposed multi-view segmentation procedure overcomes the wrongly prompting effect, by using the feature point correspondence instead of the projection method. However, these partially obstructed segmented images bring multi-view inconsistency for later NeRF training procedures. As Fig.~\ref{fig:obstruction} shows, although the target bowl indicated in Fig.~\ref{fig:obstruction} (a) has been correctly segmented in Fig.~\ref{fig:obstruction} (e), the inconsistency between Fig.~\ref{fig:obstruction} (b) and Fig.~\ref{fig:obstruction} (e) will lead to performance degradation during novel views synthesizing. 

\begin{figure}[t]
    \centering
    \includegraphics[width=\linewidth]{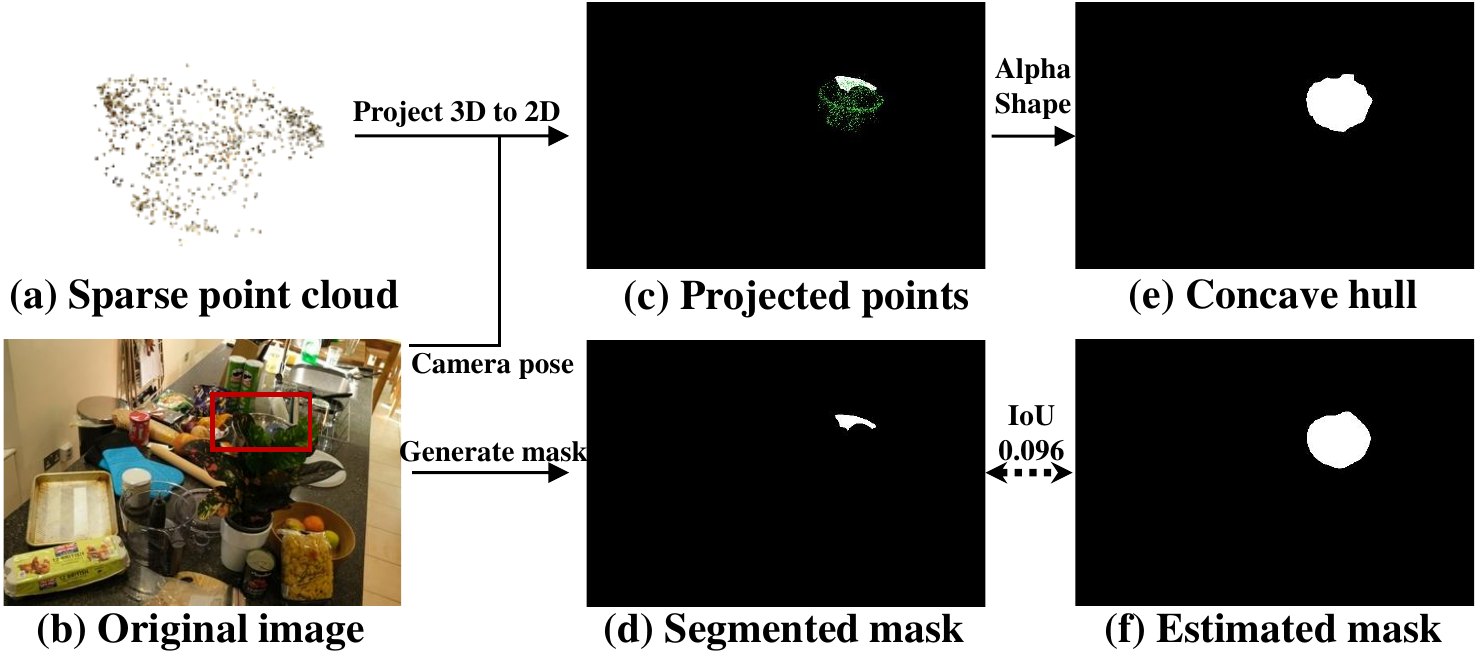}
    \caption{Procedures to identify obstructed images; 
    (a) Segmented sparse point cloud of the target object; (b) Original RGB image waiting to be masked; (c) Projecting 3D points to 2D image with known camera pose; (d) Generating mask for the original image; (e) Alpha-shape concave hull for the points; (f) Estimated mask after Gaussian filtering.
    }
    \label{fig:obstruction procedure}
\end{figure}

In order to eliminate the inconsistency brought by obstructed images, it is important to identify them first. With the segmented sparse point cloud $\bm{D}$, we can first project these 3D points to each image to get the 2D coordinates for these feature points. Then, we can construct the concave hall for these 2D points as the mask of the target object using the alpha-shape method~\cite{fischer2000introduction}. 
After that, the estimated mask can be smoothed by a Gaussian filter. Fig.~\ref{fig:obstruction procedure} shows the procedures above to identify the obstructed images. The IoU between Fig.~\ref{fig:obstruction procedure} (d) and Fig.~\ref{fig:obstruction procedure} (f) is 0.096, which means the segmented image should be discarded. It should be noted that we cannot simply calculate the convex hull and regard it as the mask, for there are usually some outlier points which will extremely affect it. 

\subsubsection{Multi-object Segmentation}

The proposed segmentation algorithm can be extended to $k$-object segmentation tasks. After giving initial prompts for each target object, we can construct $k$ 3D point lists $\bm{D}_1, \bm{D}_2, ..., \bm{D}_k$ and update them with masks separately with almost little increase in time consumption. In this way, several target objects can be segmented for only one time. The performances of the multi-object segmentation method will be shown in Subsection \ref{Results}. 


\subsection{Large Dataset Self-prompting} 
\label{Large Dataset Self-prompting}

In \cref{Multi-view Segmentation}, we propose a method which receives point prompts as input and outputs multi-view segmented views. Thus, if we can segment every object in a large dataset like ScanNet~\cite{dai2017scannet}, which contains 1000+ scenes, a large multi-view 3D object dataset can be constructed and is useful for many downstream works including generative tasks, like zero123~\cite{liu2023zero}. However, manually labeling the prompts for each object is tedious and unrealistic, which pushes us to find a feasible way to generate prompts for each scene quickly from a text prompt. 

First, a text prompt should be converted to something SAM can utilize and then generate a proper mask. Here, we use an object detector named Grounding DINO model~\cite{liu2023grounding}, which receives text input and outputs boxes and scores that indicate the position and the probability of the target object. Then, the box with the highest score can be considered as an input to SAM, which provides a proper mask for the target object. 

\begin{figure}[t]
    \centering
    \includegraphics[width=\linewidth]{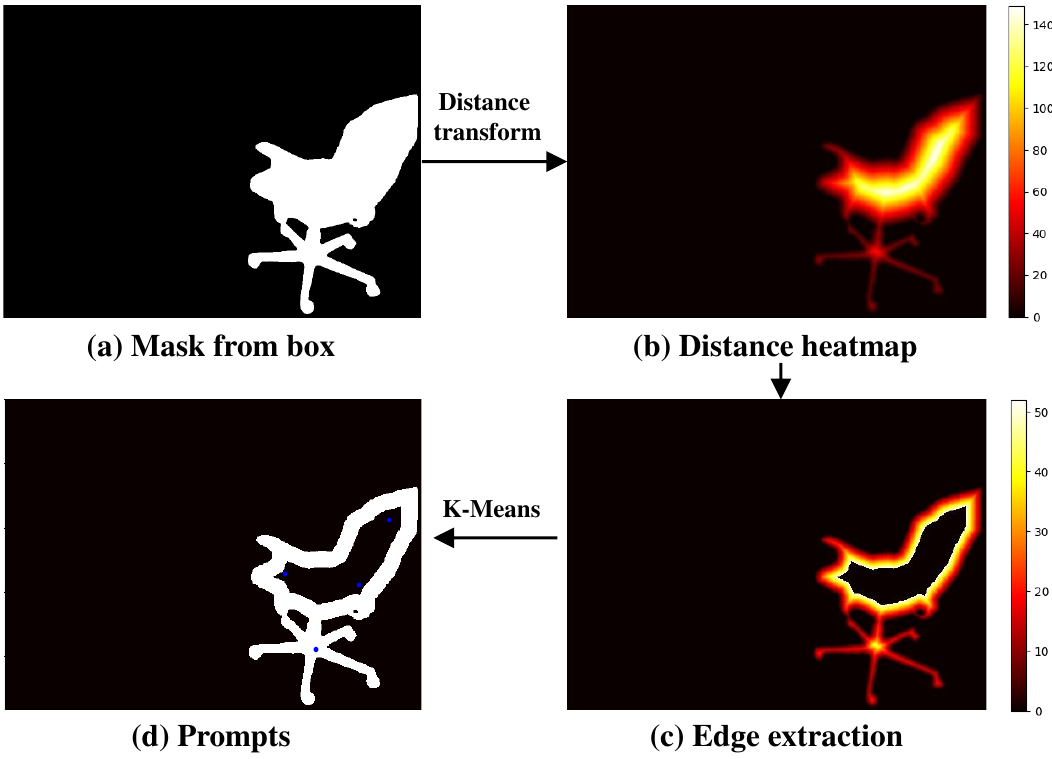}
    \caption{The procedure from mask to point prompts; (a) Mask from SAM and the box prompt; (b) Distance heatmap showing the distance to the edge for each point; (c) Extracted points which are near the edge; (d) Point prompts from $k$-means method.}
    \label{fig:generate points}
\end{figure}

The next step is generating point prompts to fit the requirements of the segmentation algorithm proposed in Subsection \ref{Multi-view Segmentation}. These point prompts should fulfill the conditions below: (1) They stay away from each other and represent all parts of the object; (2) They cannot stay too close to the edge. Thus, we can first calculate the distance to their mask for each point on the mask. Then these points near the edge are selected, for the interior points will interfere with the next step. Finally, point prompts can be generated through the $k$-means method~\cite{sinaga2020unsupervised}. Fig.~\ref{fig:generate points} shows the steps which provide point prompts from the mask. 

In this way, we create an object NeRF dataset including a large number of objects with just a few textual inputs. 
Details are discussed in \ref{Results}.


\subsection{Novel View Synthesizing} 
\label{Novel View Synthesizing} 

With the multi-view segmented images from Subsection \ref{Multi-view Segmentation}, it is practicable to synthesize novel views for the target object after training a NeRF. However, simply constructing NeRF with segmented images only will not lead to a perfect performance. In this subsection, we will introduce some methods which significantly increase the quality of synthesizing. 

\subsubsection{Sparse and Dense Depth-Supervised NeRF} 

In order to acquire better performance and faster convergence, Deng et al.~\cite{deng2022depth} have proposed a method that adds depth information to supervise the NeRF training procedure. Specifically, the segmented object sparse point cloud $\bm{D}$ mentioned in Subsection \ref{Multi-view Segmentation} will provide their 3D coordinate information. Thus, for each image, the depth of feature points corresponding to the 3D point cloud can be calculated respectively. In this way, the sparse depth supervised NeRF training can be realized with the loss as follows, 
\begin{equation}
    \mathcal{L}_{\text{NeRF}} = \mathcal{L}_{\text{rgb}} + \lambda_d \mathcal{L}_{\text{depth}}, 
\end{equation}
where $\mathcal{L}_{\text{depth}} = \Vert \bm{d} - \hat{\bm{d}} \Vert^2$ indicates the mean square error of depth. It should be noticed that sparse depth supervision makes better performance in extremely few multi-view images like less than 10. For more input images, it will also improve the quality for reconstruction 3D mesh but may not for the reconstructed RGB images~\cite{deng2022depth}. 

\begin{figure*}
    \centering
    \includegraphics[width=\linewidth]{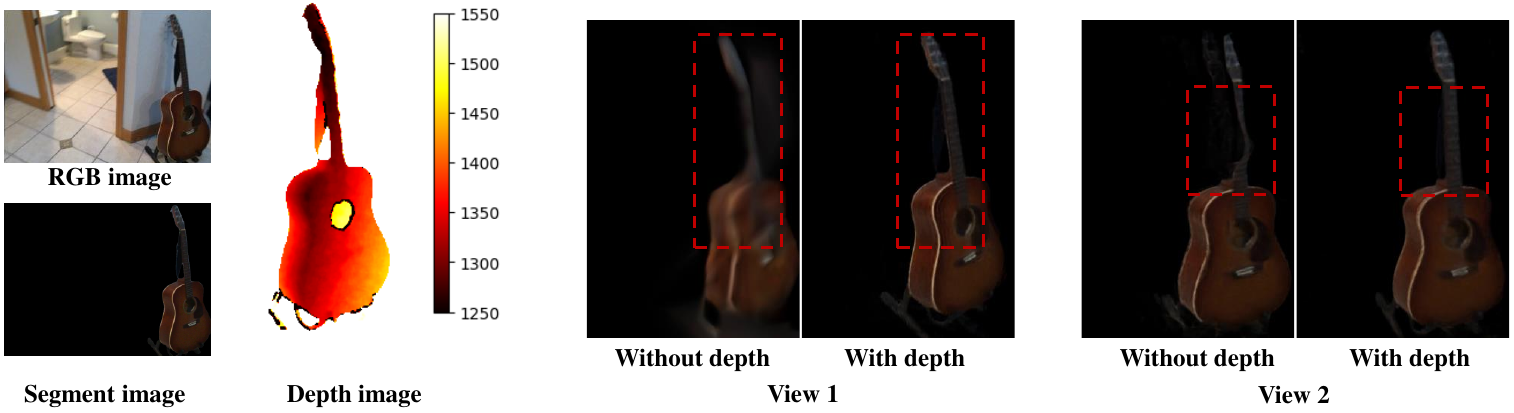}
    \caption{Comparison of novel view reconstruction performance with and without dense depth supervision.}
    \label{fig:depth supervision}
\end{figure*}

Sparse depth supervision brings limited performance enhancement due to the scarcity of depth information. To achieve higher reconstruction, dense depth information should be included in NeRF training. Many large multi-view datasets include depth image for each RGB image, such as ScanNet~\cite{dai2017scannet}, ScanNet++~\cite{yeshwanth2023scannet++}, and 3RScan~\cite{wald2019rio}, which will provide required dense depth information. Fig.~\ref{fig:depth supervision} shows the novel-view reconstruction performance comparison with and without dense depth supervision. Comparing Fig.~\ref{fig:depth supervision} (d) and Fig.~\ref{fig:depth supervision}, reconstruction with depth supervision will provide a significantly higher quality 3D mesh. 

\begin{figure}
    \centering
    \includegraphics[width=\linewidth]{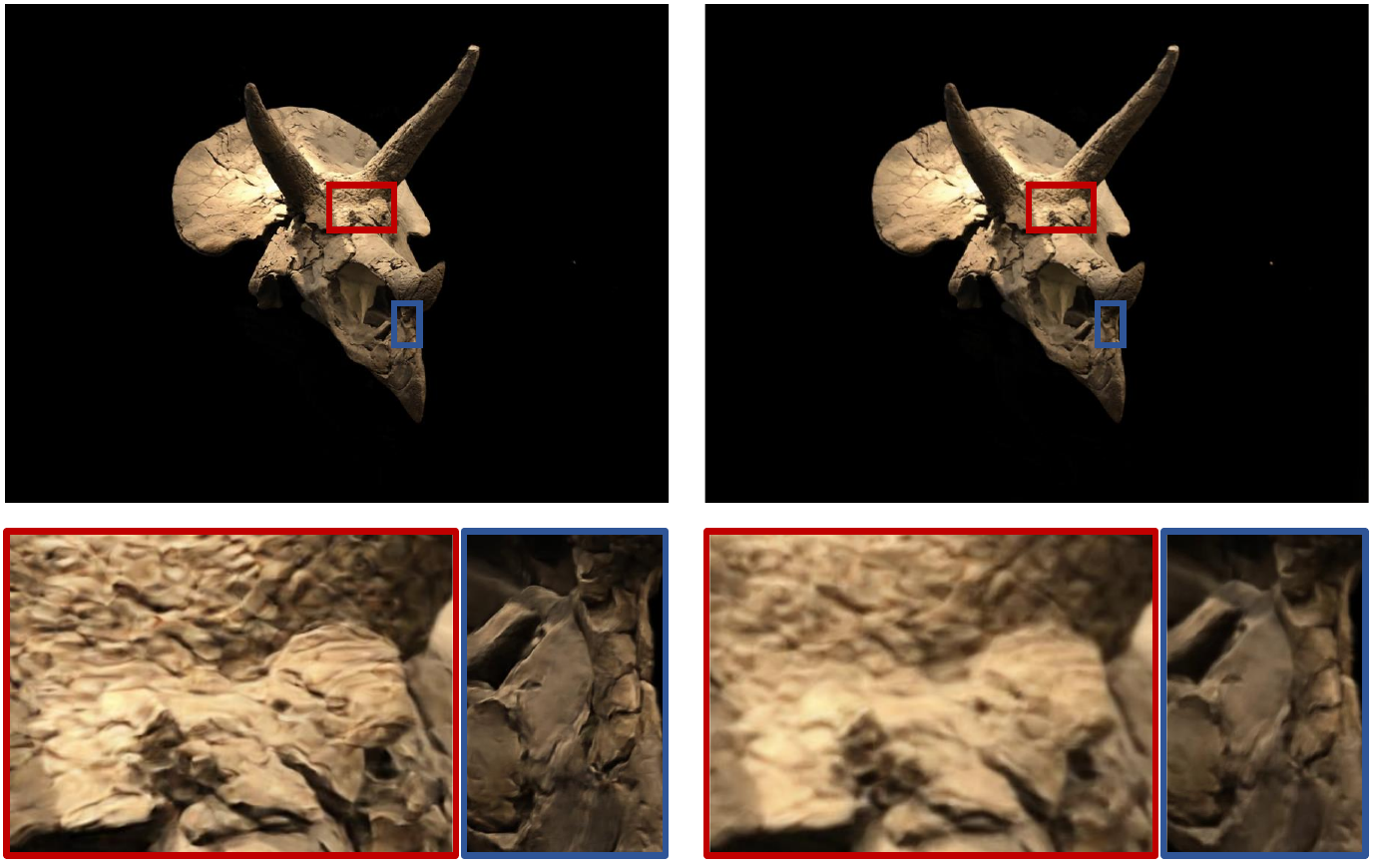}
    \caption{Comparison of reconstruction performance with different resolution training images; Left: No down-sampling with ray pruningS; Right: Down-sampling. 
    }
    \label{fig:resolution comparison}
\end{figure}

\begin{figure}[t]
    \centering
    \includegraphics[width=\linewidth]{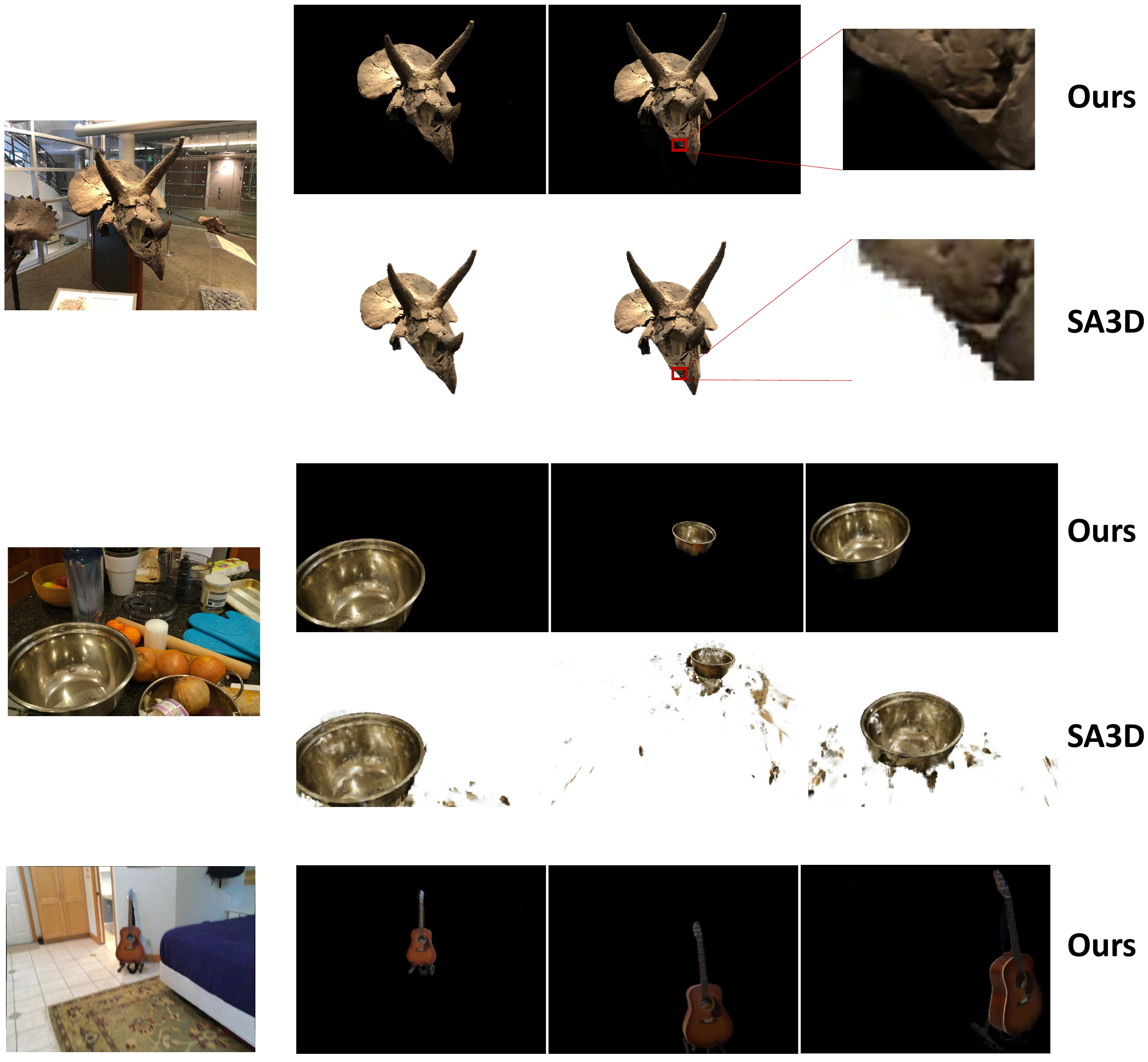}
    \caption{Novel view synthesis performance for indicated objects compared with SA3D~\cite{cen2023segment}.
    }
    \label{fig:compare to sa3d}
\end{figure}

\begin{figure*}[h]
    \centering
    \includegraphics[width=\linewidth]{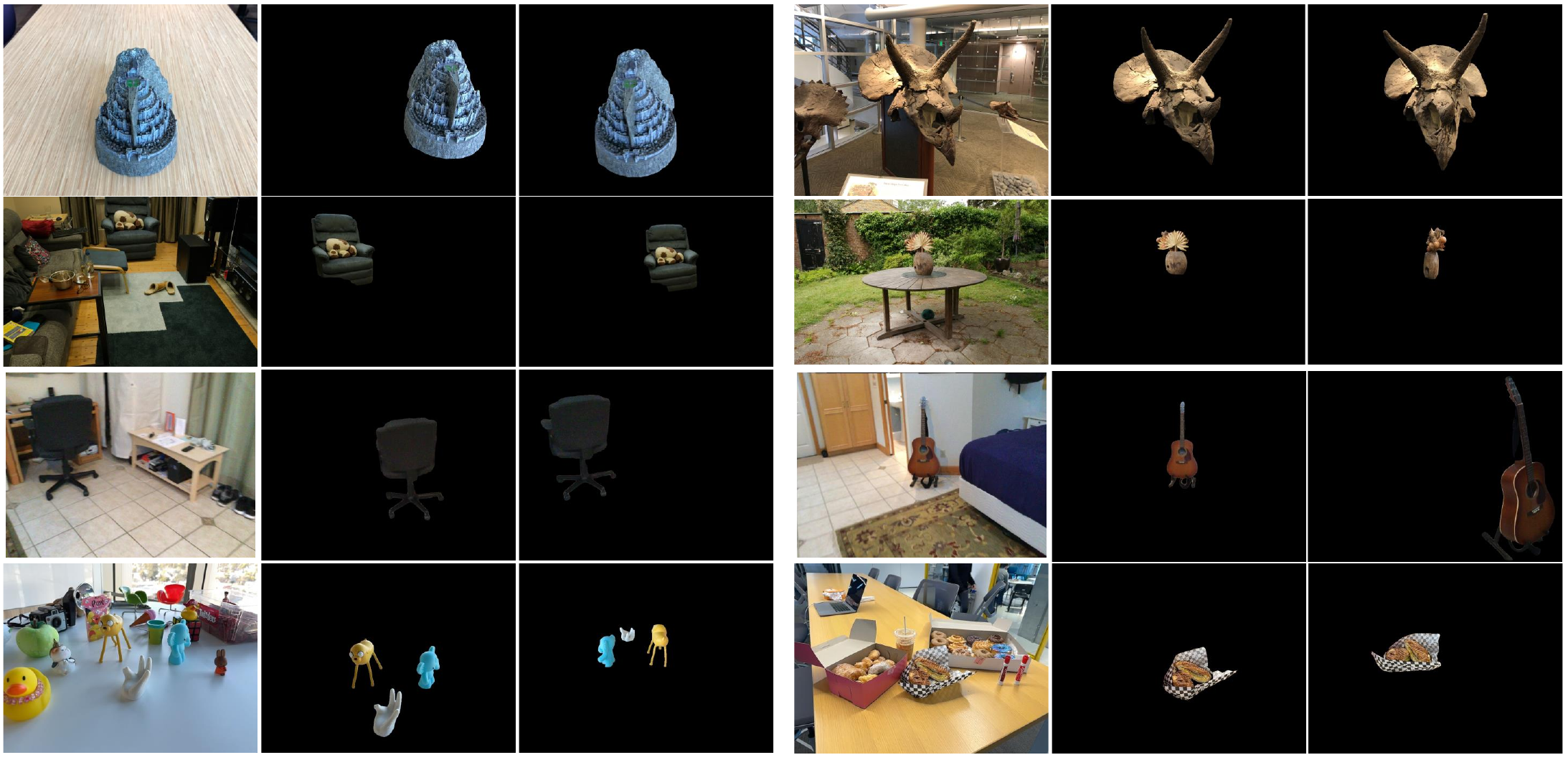}
    \caption{Performance of the multi-view segmentation procedure; First row: LLFF dataset; Second row: Mip NeRF 360 dataset; Third row: ScanNet dataset; Last row: LERF dataset.}
    \label{fig:multiview segmentation}
\end{figure*}

\begin{figure*}[h]
    \centering
    \includegraphics[width=\linewidth]{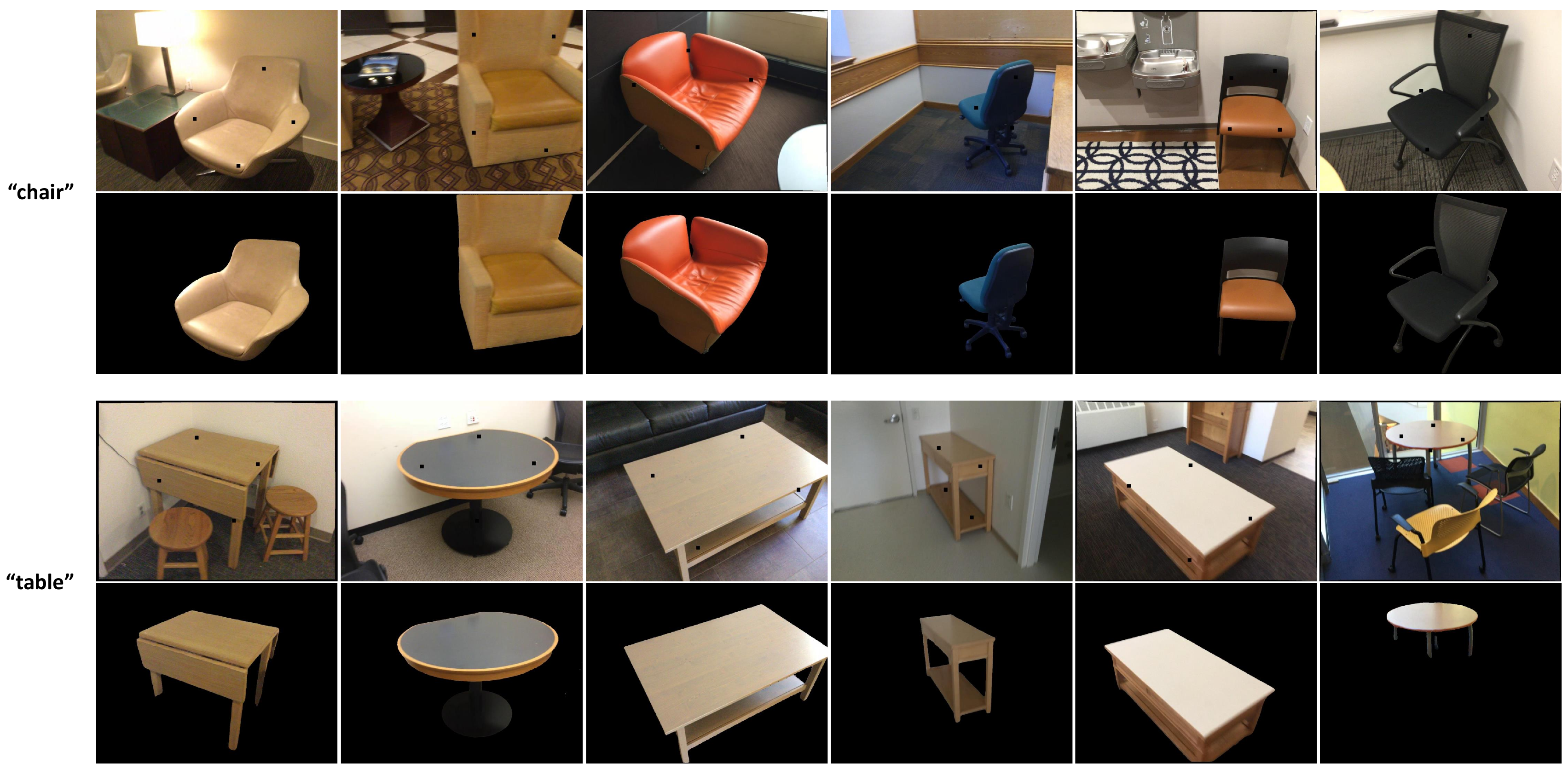}
    \caption{Construction of the multi-view objects dataset; With a textual input like "chair" or "table", the initial prompts are generated automatically for each scene.}
    \label{fig:dataset construction}
\end{figure*}

\subsubsection{Bounding Box and Ray Pruning} 
After Segmenting the indicated object from each whole image, there are three notable advantages of the reconstruction as follows: (1) Eliminate the extra components thereby reducing the additional NeRF training cost; (2) Reduce the world size leading to augmented ray sampling density and voxel density; (3) Pruning of rays unrelated to the object, significantly conserving CUDA memory and enabling the utilization of higher resolution images. In order to achieve these advantages above, some methods will be introduced during the NeRF training procedure. 

According to the segmented sparse point cloud in Subsection \ref{Multi-view Segmentation}, a bounding box $\bm{B}$ can be calculated from the known 3D point coordinates, which provides the scale of the world size in NeRF settings. With the much smaller bounding box, the density of ray sampling and the voxel grid increase accordingly (e.g. in the first column of Fig.~\ref{fig:segmented point cloud}, the overall voxel grid size decreases to 1\% of the original one). 
Additionally, any rays which not intersect with the bounding box, i.e. out of the box, will be pruned and not be used to supervise the training. In this way, the number of effective training rays is reduced by an order of magnitude, which makes the utilization of higher resolution possible. As shown in Fig.~\ref{fig:resolution comparison}, with higher resolution input RGB images, the reconstruction performance increases accordingly.

\section{Experiments} 
\label{Experiments} 

\begin{figure*}
    \centering
    \vspace{10pt}
    \includegraphics[width=\linewidth]{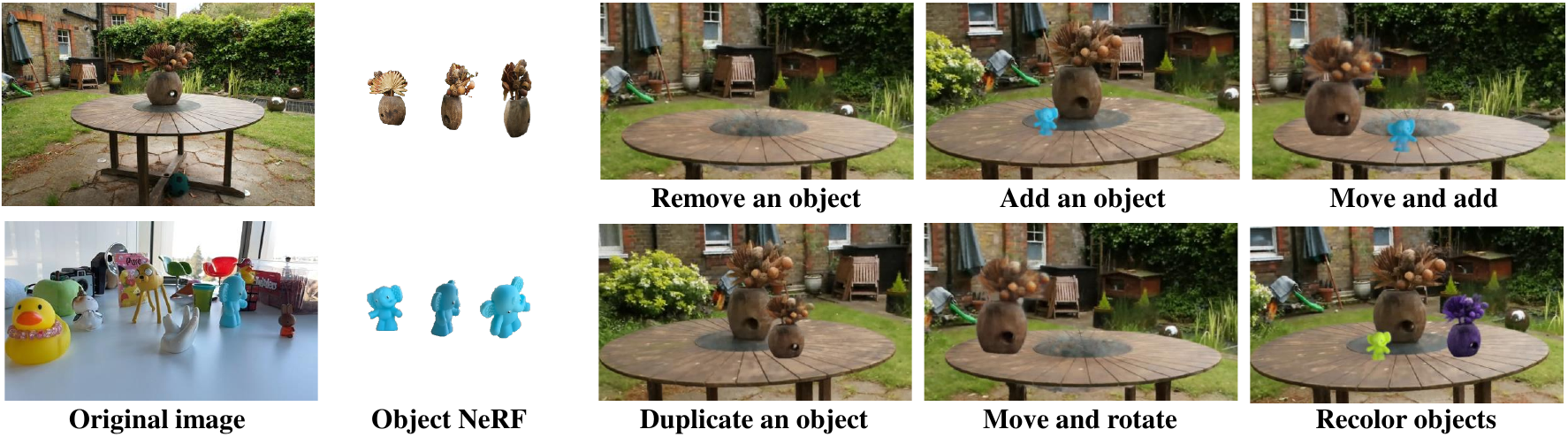}
    \caption{Applications of Obj-NeRF: editing NeRFs with object removal, add-on, movement, rotation, and color changing.}
    \label{fig:applications}
\end{figure*}

\subsection{Implementation Details} 
\label{Implementation Details} 

\textbf{Dataset.} 
In order to verify the generality of our proposed comprehensive pipeline, we evaluate the Obj-NeRF on various multi-view datasets, including face-forwarding LLFF dataset~\cite{mildenhall2019local}, Mip NeRF 360 dataset~\cite{barron2022mip}, LERF dataset~\cite{kerr2023lerf}, and large indoor datasets such as 3RScan~\cite{wald2019rio}, ScanNet~\cite{dai2017scannet}, and ScanNet++~\cite{yeshwanth2023scannet++}. Obj-NeRF will provide an indicated object NeRF with only a single prompt input for any scenario in these datasets. For large indoor datasets, the self-prompting procedure mentioned in Subsection \ref{Large Dataset Self-prompting} can be used. It will eventually provide a large multi-view object dataset including thousands of objects. 

\textbf{Novel-view Synthesis.} 
In our process of synthesizing novel views, we utilize the framework of DVGO NeRF~\cite{sun2022direct}. It is important to note that our method is not limited to DVGO, other implementations of NeRF such as Instant NGP~\cite{muller2022instant} or NeRF Studio~\cite{tancik2023nerfstudio} can also be used. Moreover, we have improved the quality of reconstruction by adopting the depth-supervision method from DS-NeRF~\cite{deng2022depth}. To achieve object NeRF applications like object removal, replacement, rotation, and color-changing, we have used Blender to generate appropriate camera poses.

\subsection{Results} 
\label{Results} 

\textbf{Multi-view Segmentation Consistency.}
As Fig.~\ref{fig:multiview segmentation} shown, the proposed multi-view segmentation algorithm demonstrates strong robustness in various datasets, including face-forwarding, ${360}^\circ$ panoramic, and large indoor scenes. Especially in the third row in Fig.~\ref{fig:multiview segmentation}, images in ScanNet dataset~\cite{dai2017scannet}
have relatively low resolution and sometimes loss of focus, our proposed procedure also works. 

\textbf{Multi-view Object Dataset.}
We utilize our proposed self-prompting method on some large indoor datasets in order to construct a multi-view object dataset. As Fig.~\ref{fig:dataset construction} has shown, after indicating a textual input like "chair" or "table", it will automatically generate initial prompts for the target object in each scene. After that, the multi-view segmentation and NeRF training procedures are followed, constructing an object NeRF for each object.

\textbf{Novel-view Synthesis.}
We construct the object NeRF under the supervision of the multi-view segmentation images mentioned above. With the methods introduced in Section \ref{Methods}, our novel view synthesis procedure overcomes the obstruction effect, enables multi-object reconstruction, and utilizes techniques that improve the reconstruction performance. As shown in Fig.~\ref{fig:compare to sa3d}, we compare our proposed method to SA3D~\cite{cen2023segment} segmenting foreground NeRF from an pre-trained full-scene NeRF,
which suffers from low resolution and floaters. Our proposed method achieves relatively high reconstruction quality across various scenarios, especially for large indoor datasets like the last row of Fig.~\ref{fig:compare to sa3d}, where the full-scene NeRF required for SA3D is impractical and low-quality.

\subsection{Applications}
\label{Applications}

In order to verify the effectiveness of the object NeRF dataset, we utilize the extracted object NeRF in various applications as shown in Fig.~\ref{fig:applications}, including object removal, replacement, rotation, and color changing. 

\textbf{Add-on.}
We can integrate the segmented object NeRF into any existing NeRF to realize the add-on task. During this process, we can also apply the rotation, resize, and other transformations to the object NeRF. Nerfstudio and blender~\cite{tancik2023nerfstudio} provides a user-friendly way to construct the required camera poses during the editing procedure. 

\textbf{Removal.} 
After obtaining the multi-view segmentation for each image, we can add a reverse alpha channel to the original image, representing the background environment without the foreground object. During the NeRF training procedure, the obstructed areas by the foreground object in one view can be inferred by other views. In this way, the object removal NeRF can be realized.

\section{Conclusions} 
\label{Conclusions}

In this paper, we propose a comprehensive pipeline for constructing segmented object NeRFs, combining the 2D segmentation proficiency of SAM and the 3D reconstruction ability of NeRF. Without dependence on full-scene NeRF, our proposed Obj-NeRF is widely applicable to various scenarios. Compared to existing works, our method outperforms on reconstruction quality and the extensiveness of application environments. Additionally, we provide a feasible way to construct a large object NeRF dataset, which is verified in some applications like NeRF editing tasks. 
For future works, the constructed object NeRF dataset can be extended to 3D generation tasks. 

{
    \small
    \bibliographystyle{ieeenat_fullname}
    \bibliography{main}

\begin{thebibliography}{35}
\providecommand{\natexlab}[1]{#1}
\providecommand{\url}[1]{\texttt{#1}}
\expandafter\ifx\csname urlstyle\endcsname\relax
  \providecommand{\doi}[1]{doi: #1}\else
  \providecommand{\doi}{doi: \begingroup \urlstyle{rm}\Url}\fi

\bibitem[Barron et~al.(2021)Barron, Mildenhall, Tancik, Hedman, Martin-Brualla,
  and Srinivasan]{barron2021mip}
Jonathan~T Barron, Ben Mildenhall, Matthew Tancik, Peter Hedman, Ricardo
  Martin-Brualla, and Pratul~P Srinivasan.
\newblock Mip-nerf: A multiscale representation for anti-aliasing neural
  radiance fields.
\newblock In \emph{Proceedings of the IEEE/CVF International Conference on
  Computer Vision}, pages 5855--5864, 2021.

\bibitem[Barron et~al.(2022)Barron, Mildenhall, Verbin, Srinivasan, and
  Hedman]{barron2022mip}
Jonathan~T Barron, Ben Mildenhall, Dor Verbin, Pratul~P Srinivasan, and Peter
  Hedman.
\newblock Mip-{NeRF} 360: Unbounded anti-aliased neural radiance fields.
\newblock In \emph{Proceedings of the IEEE/CVF Conference on Computer Vision
  and Pattern Recognition}, pages 5470--5479, 2022.

\bibitem[Carion et~al.(2020)Carion, Massa, Synnaeve, Usunier, Kirillov, and
  Zagoruyko]{carion2020end}
Nicolas Carion, Francisco Massa, Gabriel Synnaeve, Nicolas Usunier, Alexander
  Kirillov, and Sergey Zagoruyko.
\newblock End-to-end object detection with transformers.
\newblock In \emph{European Conference on Computer Vision}, pages 213--229.
  Springer, 2020.

\bibitem[Caron et~al.(2021)Caron, Touvron, Misra, J{\'e}gou, Mairal,
  Bojanowski, and Joulin]{caron2021emerging}
Mathilde Caron, Hugo Touvron, Ishan Misra, Herv{\'e} J{\'e}gou, Julien Mairal,
  Piotr Bojanowski, and Armand Joulin.
\newblock Emerging properties in self-supervised vision transformers.
\newblock In \emph{Proceedings of the IEEE/CVF International Conference on
  Computer Vision}, pages 9650--9660, 2021.

\bibitem[Cen et~al.(2023)Cen, Zhou, Fang, Shen, Xie, Zhang, and
  Tian]{cen2023segment}
Jiazhong Cen, Zanwei Zhou, Jiemin Fang, Wei Shen, Lingxi Xie, Xiaopeng Zhang,
  and Qi Tian.
\newblock Segment anything in {3D} with {NeRFs}.
\newblock \emph{arXiv preprint arXiv:2304.12308}, 2023.

\bibitem[Chen et~al.(2023)Chen, Tang, Wan, Wang, and Zeng]{chen2023interactive}
Xiaokang Chen, Jiaxiang Tang, Diwen Wan, Jingbo Wang, and Gang Zeng.
\newblock Interactive segment anything {NeRF} with feature imitation.
\newblock \emph{arXiv preprint arXiv:2305.16233}, 2023.

\bibitem[Dai et~al.(2017)Dai, Chang, Savva, Halber, Funkhouser, and
  Nie{\ss}ner]{dai2017scannet}
Angela Dai, Angel~X Chang, Manolis Savva, Maciej Halber, Thomas Funkhouser, and
  Matthias Nie{\ss}ner.
\newblock {ScanNet}: Richly-annotated {3D} reconstructions of indoor scenes.
\newblock In \emph{Proceedings of the IEEE Conference on Computer Vision and
  Pattern Recognition}, pages 5828--5839, 2017.

\bibitem[Deng et~al.(2022)Deng, Liu, Zhu, and Ramanan]{deng2022depth}
Kangle Deng, Andrew Liu, Jun-Yan Zhu, and Deva Ramanan.
\newblock Depth-supervised {NeRF}: Fewer views and faster training for free.
\newblock In \emph{Proceedings of the IEEE/CVF Conference on Computer Vision
  and Pattern Recognition}, pages 12882--12891, 2022.

\bibitem[Fischer(2000)]{fischer2000introduction}
Kaspar Fischer.
\newblock Introduction to alpha shapes.
\newblock \emph{Department of Information and Computing Sciences, Faculty of
  Science, Utrecht University}, 17, 2000.

\bibitem[Jia et~al.(2021)Jia, Yang, Xia, Chen, Parekh, Pham, Le, Sung, Li, and
  Duerig]{jia2021scaling}
Chao Jia, Yinfei Yang, Ye Xia, Yi-Ting Chen, Zarana Parekh, Hieu Pham, Quoc Le,
  Yun-Hsuan Sung, Zhen Li, and Tom Duerig.
\newblock Scaling up visual and vision-language representation learning with
  noisy text supervision.
\newblock In \emph{International Conference on Machine Learning}, pages
  4904--4916. PMLR, 2021.

\bibitem[Kerr et~al.(2023)Kerr, Kim, Goldberg, Kanazawa, and
  Tancik]{kerr2023lerf}
Justin Kerr, Chung~Min Kim, Ken Goldberg, Angjoo Kanazawa, and Matthew Tancik.
\newblock {LERF}: Language embedded radiance fields.
\newblock In \emph{Proceedings of the IEEE/CVF International Conference on
  Computer Vision}, pages 19729--19739, 2023.

\bibitem[Kirillov et~al.(2023)Kirillov, Mintun, Ravi, Mao, Rolland, Gustafson,
  Xiao, Whitehead, Berg, Lo, et~al.]{kirillov2023segment}
Alexander Kirillov, Eric Mintun, Nikhila Ravi, Hanzi Mao, Chloe Rolland, Laura
  Gustafson, Tete Xiao, Spencer Whitehead, Alexander~C Berg, Wan-Yen Lo, et~al.
\newblock Segment anything.
\newblock \emph{arXiv preprint arXiv:2304.02643}, 2023.

\bibitem[Lin et~al.(2023)Lin, Gao, Tang, Takikawa, Zeng, Huang, Kreis, Fidler,
  Liu, and Lin]{lin2023magic3d}
Chen-Hsuan Lin, Jun Gao, Luming Tang, Towaki Takikawa, Xiaohui Zeng, Xun Huang,
  Karsten Kreis, Sanja Fidler, Ming-Yu Liu, and Tsung-Yi Lin.
\newblock {Magic3D}: High-resolution text-to-3d content creation.
\newblock In \emph{Proceedings of the IEEE/CVF Conference on Computer Vision
  and Pattern Recognition}, pages 300--309, 2023.

\bibitem[Liu et~al.(2023{\natexlab{a}})Liu, Wu, Van~Hoorick, Tokmakov,
  Zakharov, and Vondrick]{liu2023zero}
Ruoshi Liu, Rundi Wu, Basile Van~Hoorick, Pavel Tokmakov, Sergey Zakharov, and
  Carl Vondrick.
\newblock Zero-1-to-3: Zero-shot one image to {3D} object.
\newblock In \emph{Proceedings of the IEEE/CVF International Conference on
  Computer Vision}, pages 9298--9309, 2023{\natexlab{a}}.

\bibitem[Liu et~al.(2023{\natexlab{b}})Liu, Zeng, Ren, Li, Zhang, Yang, Li,
  Yang, Su, Zhu, et~al.]{liu2023grounding}
Shilong Liu, Zhaoyang Zeng, Tianhe Ren, Feng Li, Hao Zhang, Jie Yang, Chunyuan
  Li, Jianwei Yang, Hang Su, Jun Zhu, et~al.
\newblock Grounding {DINO}: Marrying {DINO} with grounded pre-training for
  open-set object detection.
\newblock \emph{arXiv preprint arXiv:2303.05499}, 2023{\natexlab{b}}.

\bibitem[Mildenhall et~al.(2019)Mildenhall, Srinivasan, Ortiz-Cayon, Kalantari,
  Ramamoorthi, Ng, and Kar]{mildenhall2019local}
Ben Mildenhall, Pratul~P Srinivasan, Rodrigo Ortiz-Cayon, Nima~Khademi
  Kalantari, Ravi Ramamoorthi, Ren Ng, and Abhishek Kar.
\newblock Local light field fusion: Practical view synthesis with prescriptive
  sampling guidelines.
\newblock \emph{ACM Transactions on Graphics (TOG)}, 38\penalty0 (4):\penalty0
  1--14, 2019.

\bibitem[Mildenhall et~al.(2021)Mildenhall, Srinivasan, Tancik, Barron,
  Ramamoorthi, and Ng]{mildenhall2021nerf}
Ben Mildenhall, Pratul~P Srinivasan, Matthew Tancik, Jonathan~T Barron, Ravi
  Ramamoorthi, and Ren Ng.
\newblock {NeRF}: Representing scenes as neural radiance fields for view
  synthesis.
\newblock \emph{Communications of the ACM}, 65\penalty0 (1):\penalty0 99--106,
  2021.

\bibitem[Mirzaei et~al.(2023)Mirzaei, Aumentado-Armstrong, Derpanis, Kelly,
  Brubaker, Gilitschenski, and Levinshtein]{mirzaei2023spin}
Ashkan Mirzaei, Tristan Aumentado-Armstrong, Konstantinos~G Derpanis, Jonathan
  Kelly, Marcus~A Brubaker, Igor Gilitschenski, and Alex Levinshtein.
\newblock {SPIn-NeRF}: Multiview segmentation and perceptual inpainting with
  neural radiance fields.
\newblock In \emph{Proceedings of the IEEE/CVF Conference on Computer Vision
  and Pattern Recognition}, pages 20669--20679, 2023.

\bibitem[M{\"u}ller et~al.(2022)M{\"u}ller, Evans, Schied, and
  Keller]{muller2022instant}
Thomas M{\"u}ller, Alex Evans, Christoph Schied, and Alexander Keller.
\newblock Instant neural graphics primitives with a multiresolution hash
  encoding.
\newblock \emph{ACM Transactions on Graphics (ToG)}, 41\penalty0 (4):\penalty0
  1--15, 2022.

\bibitem[Munkberg et~al.(2022)Munkberg, Hasselgren, Shen, Gao, Chen, Evans,
  M{\"u}ller, and Fidler]{munkberg2022extracting}
Jacob Munkberg, Jon Hasselgren, Tianchang Shen, Jun Gao, Wenzheng Chen, Alex
  Evans, Thomas M{\"u}ller, and Sanja Fidler.
\newblock Extracting triangular 3d models, materials, and lighting from images.
\newblock In \emph{Proceedings of the IEEE/CVF Conference on Computer Vision
  and Pattern Recognition}, pages 8280--8290, 2022.

\bibitem[Poole et~al.(2022)Poole, Jain, Barron, and
  Mildenhall]{poole2022dreamfusion}
Ben Poole, Ajay Jain, Jonathan~T Barron, and Ben Mildenhall.
\newblock Dreamfusion: Text-to-3d using 2d diffusion.
\newblock \emph{arXiv preprint arXiv:2209.14988}, 2022.

\bibitem[Pumarola et~al.(2021)Pumarola, Corona, Pons-Moll, and
  Moreno-Noguer]{pumarola2021d}
Albert Pumarola, Enric Corona, Gerard Pons-Moll, and Francesc Moreno-Noguer.
\newblock {D-NeRF}: Neural radiance fields for dynamic scenes.
\newblock In \emph{Proceedings of the IEEE/CVF Conference on Computer Vision
  and Pattern Recognition}, pages 10318--10327, 2021.

\bibitem[Radford et~al.(2021)Radford, Kim, Hallacy, Ramesh, Goh, Agarwal,
  Sastry, Askell, Mishkin, Clark, et~al.]{radford2021learning}
Alec Radford, Jong~Wook Kim, Chris Hallacy, Aditya Ramesh, Gabriel Goh,
  Sandhini Agarwal, Girish Sastry, Amanda Askell, Pamela Mishkin, Jack Clark,
  et~al.
\newblock Learning transferable visual models from natural language
  supervision.
\newblock In \emph{Proceedings of International Conference on Machine
  Learning}, pages 8748--8763. PMLR, 2021.

\bibitem[Roessle et~al.(2022)Roessle, Barron, Mildenhall, Srinivasan, and
  Nie{\ss}ner]{roessle2022dense}
Barbara Roessle, Jonathan~T Barron, Ben Mildenhall, Pratul~P Srinivasan, and
  Matthias Nie{\ss}ner.
\newblock Dense depth priors for neural radiance fields from sparse input
  views.
\newblock In \emph{Proceedings of the IEEE/CVF Conference on Computer Vision
  and Pattern Recognition}, pages 12892--12901, 2022.

\bibitem[Schonberger and Frahm(2016)]{schonberger2016structure}
Johannes~L Schonberger and Jan-Michael Frahm.
\newblock Structure-from-motion revisited.
\newblock In \emph{Proceedings of the IEEE Conference on Computer Vision and
  Pattern Recognition}, pages 4104--4113, 2016.

\bibitem[Sinaga and Yang(2020)]{sinaga2020unsupervised}
Kristina~P Sinaga and Miin-Shen Yang.
\newblock Unsupervised k-means clustering algorithm.
\newblock \emph{IEEE Access}, 8:\penalty0 80716--80727, 2020.

\bibitem[Sun et~al.(2022)Sun, Sun, and Chen]{sun2022direct}
Cheng Sun, Min Sun, and Hwann-Tzong Chen.
\newblock Direct voxel grid optimization: Super-fast convergence for radiance
  fields reconstruction.
\newblock In \emph{Proceedings of the IEEE/CVF Conference on Computer Vision
  and Pattern Recognition}, pages 5459--5469, 2022.

\bibitem[Tancik et~al.(2023)Tancik, Weber, Ng, Li, Yi, Wang, Kristoffersen,
  Austin, Salahi, Ahuja, et~al.]{tancik2023nerfstudio}
Matthew Tancik, Ethan Weber, Evonne Ng, Ruilong Li, Brent Yi, Terrance Wang,
  Alexander Kristoffersen, Jake Austin, Kamyar Salahi, Abhik Ahuja, et~al.
\newblock Nerfstudio: A modular framework for neural radiance field
  development.
\newblock In \emph{ACM SIGGRAPH 2023 Conference Proceedings}, pages 1--12,
  2023.

\bibitem[Tang et~al.(2022)Tang, Chen, Wang, and Zeng]{tang2022compressible}
Jiaxiang Tang, Xiaokang Chen, Jingbo Wang, and Gang Zeng.
\newblock Compressible-composable {NeRF} via rank-residual decomposition.
\newblock \emph{Advances in Neural Information Processing Systems},
  35:\penalty0 14798--14809, 2022.

\bibitem[Wald et~al.(2019)Wald, Avetisyan, Navab, Tombari, and
  Nie{\ss}ner]{wald2019rio}
Johanna Wald, Armen Avetisyan, Nassir Navab, Federico Tombari, and Matthias
  Nie{\ss}ner.
\newblock {RIO}: {3D} object instance re-localization in changing indoor
  environments.
\newblock In \emph{Proceedings of the IEEE/CVF International Conference on
  Computer Vision}, pages 7658--7667, 2019.

\bibitem[Wang et~al.(2022)Wang, Wu, Guo, Zhang, Tai, and Hu]{wang2022nerf}
Chen Wang, Xian Wu, Yuan-Chen Guo, Song-Hai Zhang, Yu-Wing Tai, and Shi-Min Hu.
\newblock {NeRF-SR}: High quality neural radiance fields using supersampling.
\newblock In \emph{Proceedings of the 30th ACM International Conference on
  Multimedia}, pages 6445--6454, 2022.

\bibitem[Yeshwanth et~al.(2023)Yeshwanth, Liu, Nie{\ss}ner, and
  Dai]{yeshwanth2023scannet++}
Chandan Yeshwanth, Yueh-Cheng Liu, Matthias Nie{\ss}ner, and Angela Dai.
\newblock {ScanNet++}: A high-fidelity dataset of {3D} indoor scenes.
\newblock In \emph{Proceedings of the IEEE/CVF International Conference on
  Computer Vision}, pages 12--22, 2023.

\bibitem[Yin et~al.(2023)Yin, Fu, Yang, and Lin]{yin2023or}
Youtan Yin, Zhoujie Fu, Fan Yang, and Guosheng Lin.
\newblock {OR-NeRF}: Object removing from {3D} scenes guided by multiview
  segmentation with neural radiance fields.
\newblock \emph{arXiv preprint arXiv:2305.10503}, 2023.

\bibitem[Yuan et~al.(2022)Yuan, Sun, Lai, Ma, Jia, and Gao]{yuan2022nerf}
Yu-Jie Yuan, Yang-Tian Sun, Yu-Kun Lai, Yuewen Ma, Rongfei Jia, and Lin Gao.
\newblock {NeRF}-editing: geometry editing of neural radiance fields.
\newblock In \emph{Proceedings of the IEEE/CVF Conference on Computer Vision
  and Pattern Recognition}, pages 18353--18364, 2022.

\bibitem[Zou et~al.(2023)Zou, Yang, Zhang, Li, Li, Gao, and
  Lee]{zou2023segment}
Xueyan Zou, Jianwei Yang, Hao Zhang, Feng Li, Linjie Li, Jianfeng Gao, and
  Yong~Jae Lee.
\newblock Segment everything everywhere all at once.
\newblock \emph{arXiv preprint arXiv:2304.06718}, 2023.

\end{thebibliography}
}


\end{document}